# Automated Skin Lesion Classification Using Ensemble of Deep Neural Networks in ISIC 2018: Skin Lesion Analysis Towards Melanoma Detection Challenge

Md Ashraful Alam Milton

Auonomous Unversity of Barcelona

Campus de la UAB, Plaça Cívica, 08193 Bellaterra, Barcelona, Spain

mdashrafulalam.milton@e-campus.uab.cat

## Abstract

*In this paper, we studied extensively on different deep learning based methods to detect melanoma and skin lesion cancers. Melanoma, a form of malignant skin cancer is very threatening to health. Proper diagnosis of melanoma at an earlier stage is crucial for the success rate of complete cure. Dermoscopic images with Benign and malignant forms of skin cancer can be analyzed by computer vision system to streamline the process of skin cancer detection. In this study, we experimented with various neural networks which employ recent deep learning based models like PNASNet-5-Large, InceptionResNetV2, SENet154, InceptionV4. Dermoscopic images are properly processed and augmented before feeding them into the network. We tested our methods on International Skin Imaging Collaboration (ISIC) 2018 challenge dataset. Our system has achieved best validation score of 0.76 for PNASNet-5-Large model. Further improvement and optimization of the proposed methods with a bigger training dataset and carefully chosen hyper-parameter could improve the performances. The code available for download at https://github.com/miltonbd/ISIC_2018_classification.*

## 1. Introduction

Skin cancer is the uncontrolled growth of abnormal skin cells. It occurs when DNA damage to skin cells (most often caused by ultraviolet radiation from sunshine or tanning beds) triggers mutations, or genetic defects, that lead the skin cells to multiply rapidly and form malignant tumors. Except in rare instances, most skin cancers arise from DNA mutations induced by ultraviolet light affecting cells of the epidermis.

Many of these early cancers seem to be controlled by natural immune surveillance, which when compromised, may permit the development of masses of malignant cells that begin to grow into tumors. Cancer of the skin is by far the most common of all cancers. Melanoma accounts for only about 1% of skin cancers but causes a large majority of skin cancer deaths.

The American Cancer Society estimates for melanoma[4] in the United States for 2018 are: About 91,270 new melanomas will be diagnosed (about 55,150 in men and 36,120 in women). About 9,320 people are expected to die of melanoma (about 5,990 men and 3,330 women). The rates of melanoma have been rising for the last 30 years. Melanoma is more than 20 times more common in whites than in African Americans. Overall, the lifetime risk of getting melanoma is about 2.6% (1 in 38) for whites, 0.1% (1 in 1,000) for blacks, and 0.58% (1 in 172) for Hispanics.

Skin cancer is traditionally diagnosed by physical examination and biopsy. The biopsy is a quick and simple procedure where part or all of the spot is removed and sent to a laboratory. It may be done by your doctor or you might be referred to a dermatologist or surgeon. Results may take about a week to come through. This manual diagnosis is time-consuming, expensive and may produce the wrong result due to the bias of the dermatologist.Recent years, we have seen deep learning based classification system surpasses human in big datasets like Imagenet[1], MSCOCO[18], and PASCAL[17]. Automated skin lesion classification can be a viable approach to deal with the diagnosis of melanoma.

## 2. Background And Related work

Using digital image processing in an attempt to classify benign and malignant skin lesions from dermoscopy images[2]. Deep neural networks achieved dermatologists level accuracy[3]. Dermatologists have devised several techniques to classify lesions.

Three-point checklist is a typical method for diagnosis of melanoma and skin lesion is considered to be melanoma if at-least two of them are true[4]:

• Asymmetry: asymmetry of color and structure in one or two perpendicular axes

• Atypical network: pigment network with irregular holes and thick lines

• Blue-white structures: any type of blue and/or white color, i.e. combination of blue-whiteveil and regression structures.

ABCD parameters method is also common for diagnosis of melanoma[5]:

• Asymmetrical shape - melanoma lesions are typically asymmetrical

• Borders - melanoma lesions have the irregular border

• Color - the presence of more than one color in melanoma lesions

• Diameter - melanoma lesions are typically larger than 6mm in diameterThe work of Kawahara et al. [9] explores the idea of using a pre-trained ConvNet as a feature extractor rather than training a CNN from scratch.

Furthermore, it demonstrates the use of filters from a CNN trained on natural images generalize to classifying 10 classes of non-dermoscopic skin images. Liao's [10] work attempted to construct a universal skin disease classification by applying transfer learning on a deep CNN and fine-tuned its weights by continuing the backpropagation.

In Codella et al. [8], the authors report new state-of-the-art performance using ConvNets to extract image descriptors by using a pre-trained model from the Image Large Scale Visual Recognition Challenge (ILSVRC) 2012 dataset [7]. They also investigate the most recent network structure to win the ImageNet recognition challenge called Deep Residual Network (DRN) [11].

## 3. ISIC 2018 Melanoma Detection Challenege and Dataset

ISIC 2018: Skin Lesion Analysis Towards Melanoma Detection has three tasks: Task 1- Lesion Segmentation, Task 2- Lesion Attribute Detection, Task 3- Disease Classification.

The dataset for workshop ISIC 2018: Skin Lesion Analysis Towards Melanoma Detection 1 is used [21], [22]. In the training set, there are a total of 10015 skin lesion images from seven skin diseases- Melanoma (1113), Melanocytic nevus (6705), Basal cell carcinoma (514), Actinic keratosis (327), Benign keratosis (1099), Dermatofibroma (115) and Vascular (142). The validation dataset consists of 193 images. Figure 1 shows some example of these 7 types of the skin lesion. Task- 3: the goal of task-3 is to find improved automated predictions of disease classification within dermoscopic images. Possible disease categories are in below figure 1.

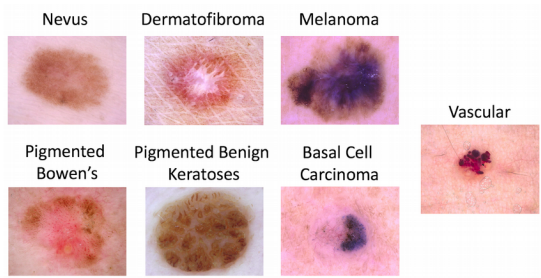

Figure 1. Example of ISIC 2018: Skin Lesion Analysis Towards Melanoma Detection dataset.

## 4. Proposed Methodology

**Pre-Processing**: The images have different sizes in the challenege. We applied three preprocessing steps on the images. First, we normalized the images by subtracting the mean RGB value of ImageNet dataset as suggested in [1]. input pixel range was converted to 1-0. After that, all images resized to appropriate size to feed into the neural network.

**Data Augmentation**: When large imbalances of the class present, minority oversampling is a commonly used technique to recover the robustness of the model and lessen the bias of the dataset. Deep learning based models are data hungry and generalize well when they are feed with a lot of data. As the training set only contains 2000 images, so we augmented these images by rotation, flip, random crop, adjust_brightness, adjust contrast, pixel jitter, Aspect Ratio, random shear, zoom, and vertical and horizontal shift and flip. This makes the training dataset less imbalanced and improves the neural network accuracy. We also used images from HAM 1000 for training.

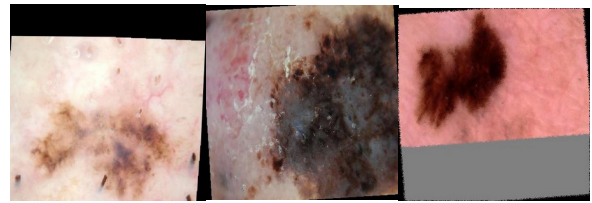

Figure 2. Augmented Images

**Representation Learning**: The goal of representation learning or feature learning is to find an appropriate representation of data in order to perform a machine learning task. In particular, deep learning exploits this concept by its very nature. In a neural network, each hidden layer maps its input data to an inner representation that tends to capture a higher level of abstraction. These learned features are increasingly more informative through layers towards the machine learning task that we intend to perform (e.g. classification).

Basically, representation learning is nothing more than a set of features that would describe concepts individually. We could even have the representation of objects using their colors, shape and size and their feature. We could even represent images based on some values.

Representation is what would help us differentiate between different concepts, and in turn, would also help us find out similarities between them. There are various methods in deep learning which helps us to extract these features and represent the concepts so that further machine learning tasks can be carried out. Representation of words

226

as vectors is a very good example of representation learning.

**Models**: We made a set of models to be used in the model ensemble including PNASNet-5-Large, InceptionResNetV2, SENet154, InceptionV4, and an ensemble of all models. These networks were used as a feature extractor to learn the internal representation of the image. As the training items are only 10k, the model trainable params should be lower. We freeze all layers except the last layer.

In the last FC layer, we also used softmax to output the probability of each category as a classification. After the standard model, we used Fully Connected Layer (FC). final output layer has 7 classes for TASK-3.

**Fine Tuning:** As the amounts of training examples is insufficient for training a deep convolutional network from scratch, so we used the pre-trained imagenet model to initialize the network parameters and fine-tuned the last several layers. We keep the weight of all layers except FC and output layer frozen in the first few iterations of training due to high unstable gradient flow. After a few iterations of training, we unfreeze the last few layers and adjust their weight by back-propagation. Fine-tuning helps to train easily and prevents it from over-fitting. We plotted the training curves in figure 3. The left one is training this network from scratch, the right one is fine-tuning, obviously, the fine-tuned network can lead to a better convergence. In our experiments, we used Pytorch[19], a deep learning library to train our models.

**Classification**: The final layer outputs a vector of which indicates the class number. The final class is one of the Melanoma, Melanocytic nevus, Basal cell carcinoma, Actinic keratosis / Bowen's disease (intraepithelial carcinoma), Benign keratosis (solar lentigo / seborrheic keratosis/lichen planus-like keratosis), Dermatofibroma, Vascular lesion.

**Evaluation**: To evaluate the classification results, we feed the validation images to our trained network and created appropriate submission file utilizing output o the network.

## 5. Training Details

For every model, we started with their imagenet pre-trained model. For the first 2 epochs, we freeze the whole network weights except the last layer and used a low learning rate of 0.0001. This due to the fact that, when fine-tuning a model in the new dataset, for the first few epochs the error gradient is very high and that may update the model in the wrong direction. After 2 epochs, we set the learning rate to 0.01 and updated all the weights of the model. We trained for 50 epoch and for each epoch we checked the validation score in the evaluation server for validation set. We employed early stopping by observing the validation error. Optimization method was Adaptive momentum (Adam), the cross-entropy loss for updating the weights by back-propagating the error backward.

## 6. Implementation Details

Both train and fine-tune actions are very time-consuming procedures, so efficient implementations are highly welcome because of the limited computational resources and the vast amount of training data. Currently, Tensorflow[20], Pytorch[19] are popular with the researcher. We have found that pytorch is highly suitable for the project due to its simplicity and flexibility in modifying the dynamic computation graph. we implemented our project in pytorch. We utilized multi gpu Nvidia 1080ti for training the networks. 1080ti each has CUDA Cores 3584, 11 GB GDDR5X, Memory Interface Width 352-bit, Memory Bandwidth (GB/sec) 11 Gbps. We used pytorch, tensorboard, scipy, numpy.

## 7. Results

| Serial No | Method | Validation Score |
|---|---|---|
| 1. | PNASNet-5-Large | 0.76 |
| 2. | InceptionResNetV2 | 0.70 |
| 3. | SENet154 | 0.74 |
| 4. | InceptionV4 | 0.67 |
| 5. | Ensemble | 0.73 |

## 8. Conclusion

We demonstrated that an ensemble of deep neural networks based methods can achieve competitive classification performance dermoscopic image classification to detect skin lesion. This could emerge as a highly automated and accurate dermoscopic image classification system that could be used along with experienced dermatologists.

Due to the unbalanced dataset with a large difference in total images for each class makes it harder to generalize the visual features of the lesions. The more balanced dataset will tend to deliver an improved result. In addition to that, the bigger dataset with improved feature variety will improve the overall performance of the classification by learning better representation, reducing the risk of over-fitting; and thus, generalize well. Moreover, performing additional regularization tweaks and fine-tuning of hyper-parameters may improve the model's robustness. The proposed approach has best validation score for PNASNet-5-Large model which is 0.76.

**ACKNOWLEDGMENT**

We are thankful to the organizer of "MICCAI 2018

challenge ISIC for providing the skin images. Automated Skin Lesion Classification Using Ensemble of Deep Neural Networks in ISIC 2018 Skin Lesion Detection Challenge.

**References**

[1] Krizhevsky, A., I. Sutskever, and G.E. Hinton. "Imagenet classification with deep convolutional neural networks". in Advances in neural information processing systems. 2012.

[2] M Emre Celebi, Teresa Mendonca, and Jorge S Marques. Dermoscopy image analysis. CRC Press, 2015.

[3] Andre Esteva, Brett Kuprel, Roberto A Novoa, Justin Ko, Susan M Swetter, Helen M Blau, and Sebastian Thrun. Dermatologist-level classification of skin cancer with deep neural networks. Nature, 542(7639):115–118, 2017.

[4] https://www.cancer.org/cancer/melanoma-skin-cancer/about/key-statistics.html, accessed in 17.05.2018

[5] H Peter Soyer, Giuseppe Argenziano, Iris Zalaudek, Rosamaria Corona, Francesco Sera, Renato Talamini, Filomena Barbato, Adone Baroni, Lorenza Cicale, Alessandro Di Stefani, et al. Three point checklist of dermoscopy. Dermatology, 208(1):27–31, 2004.

[6] K. Simonyan and A. Zisserman, "Very deep convolutional networks for large scale image recognition," arXiv preprint arXiv:1409.1556, 2014.

[7] O. Russakovsky, J. Deng, H. Su, J. Krause, S. Satheesh, S. Ma, Z. Huang, A. Karpathy, A. Khosla, M. Bernstein et al., "Imagenet large scale visual recognition challenge," International Journal of Computer Vision, vol. 115, no. 3, pp. 211–252, 2015.

[8] N. Codella, Q.-B. Nguyen, S. Pankanti, D. Gutman, B. Helba, A. Halpern, and J. R. Smith, "Deep learning ensembles for melanoma recognition in dermoscopy images," arXiv preprint arXiv:1610.04662, 2016.

[9] J. Kawahara, A. BenTaieb, and G. Hamarneh, "Deep features to classify skin lesions," IEEE International Symposium on Biomedical Imaging (IEEE ISBI), pp. 1397–1400

[10] H. Liao, "A deep learning approach to universal skin disease classification," https://www.cs.rochester.edu/u/hliao6/projects/other/skin project report.pdf.

[11] K. He, X. Zhang, S. Ren, and J. Sun, "Deep residual learning for image recognition," arXiv preprint arXiv:1512.03385, 2015. "Dermnet - skin disease atlas,"

[12] http://www.dermnet.com/ accessed in 17.05.2018.

[13] Diederik Kingma and Jimmy Ba. Adam: A method for stochastic optimization. In The International Conference on Learning Representations (ICLR), 2015.

[14] https://www.skincancer.org/skin-cancer-information accessed in 17.05.2018

[15] https://www.medicinenet.com/skin_cancer_overview/article.htm#what_are_the_different_types_of_skin_cancer

[16] https://www.quora.com/What-is-representation-learning-in-deep-learning

[17] M. Everingham, L. Van Gool, C. K. I. Williams, J. Winn, and A. Zisserman. The PASCAL Visual Object Classes Challenge 2012 (VOC2012) Results. http://www.pascal-network.org/challenges/VOC/voc2012/workshop/index.html.

[18] Tsung-Yi Lin, Michael Maire, Serge Belongie, Lubomir Bourdev, Ross Girshick, James Hays, Pietro Perona, Deva Ramanan, C. Lawrence Zitnick, Piotr Dollár, Microsoft COCO: Common Objects in Context, https://arxiv.org/abs/1405.0312

[19] Pytorch, https://pytorch.org/

[20] Tensorflow, https://www.tensorflow.org/

[21] Noel C. F. Codella, David Gutman, M. Emre Celebi, Brian Helba, Michael A. Marchetti, Stephen W. Dusza, Aadi Kalloo, Konstantinos Liopyris, Nabin Mishra, Harald Kittler, Allan Halpern: "Skin Lesion Analysis Toward Melanoma Detection: A Challenge at the 2017 International Symposium on Biomedical Imaging (ISBI), Hosted by the International Skin Imaging Collaboration (ISIC)", 2017; arXiv:1710.05006.

[22] Philipp Tschandl, Cliff Rosendahl, Harald Kittler: "The HAM10000 Dataset: A Large Collection of Multi-Source Dermatoscopic Images of Common Pigmented Skin Lesions", 2018; arXiv:1803.10417.